\documentclass[journal]{IEEEtran}
\usepackage{cite}
\usepackage{amsmath,amssymb,amsfonts}
\usepackage{algorithmic}
\usepackage{graphicx}
\usepackage[dvipsnames]{xcolor}
\usepackage{textcomp}
\usepackage{multirow}
\usepackage{vcell}
\usepackage{hhline}
\usepackage{array}
\usepackage{booktabs}

\newcommand{\bluefunc}[1]{#1}

\usepackage{ifpdf}
\ifCLASSINFOpdf
\fi

\def\BibTeX{{\rm B\kern-.05em{\sc i\kern-.025em b}\kern-.08em
    T\kern-.1667em\lower.7ex\hbox{E}\kern-.125emX}}

\usepackage{fancyhdr}

\usepackage{dblfloatfix}

\begin{document}

\title{AdvMT: Adversarial Motion Transformer for Long-term Human Motion Prediction}

\author{Sarmad~Idrees\textsuperscript{1}, Seokman~Sohn\textsuperscript{2}, and~Jongeun~Choi\textsuperscript{1,*}%
\thanks{\textsuperscript{1} Sarmad Idrees and Jongeun Choi are with the School of Mechanical Engineering, Yonsei University, 50 Yonsei-Ro, Seodaemun-Gu, Seoul 03722, South Korea (e-mail: sarmad@yonsei.ac.kr). \textsuperscript{*}Correspondence to Jongeun Choi (jongeunchoi@yonsei.ac.kr).}%
\thanks{\textsuperscript{2} Seokman Sohn is with the Power Generation Lab, Korea Electric Power Research Institute, Munji-Ro, Yuseong-Gu, Daejeon 34056, South Korea (e-mail: happysohn@kepco.co.kr).}%
\thanks{This work was supported by Korea Institute of Energy Tech-nology Evaluation and Planning (KETEP) grant funded by the Korea government (MOTIE) (20206610100290, Development of Work Safe-ty  Management  Platform  in  Power  Plant).  This  work  was  supported  in  part  by  the  National  Research  Foundation  of  Korea  (NRF)  grants  funded by the Korea government (MSIT) (No.RS-2024-00344732).}%
}

% \markboth{IEEE ROBOTICS AND AUTOMATION LETTERS}%
% {Idrees \MakeLowercase{\textit{et al.}}: AdvMT: Adversarial Motion Transformer for Long-term Human Motion Prediction}

\maketitle

\begin{abstract}
Human motion prediction is a crucial capability for advanced robotic systems that interact with humans. In facilities with dynamic human-robot collaboration settings, robots must anticipate human movements to ensure safety, prevent collisions, and optimize cooperative tasks. Traditionally, motion forecasting is treated as a sequential modeling problem using historical pose data, but achieving long-term accuracy and physical realism remains challenging. We present Adversarial Motion Transformer (AdvMT), a novel approach that integrates a Transformer-based motion encoder with a temporal continuity discriminator to address these challenges. The Transformer captures rich spatio-temporal dependencies across human joints, while adversarial training with a continuity discriminator enforces smooth, natural motion trajectories that adhere to biomechanical constraints. Our training scheme includes a bone-length consistency term and adversarial loss to reduce common artifacts like pose freezing or unnatural transitions. In experiments on the Human3.6M motion dataset, AdvMT achieves state-of-the-art long-horizon prediction accuracy while also delivering robust short-term predictions. \bluefunc{These improvements strengthen the prediction foundation for physical AI in manufacturing and human-robot collaboration, where anticipating human motion is a prerequisite for safe and efficient robot coordination.}
\end{abstract}


% \fi
%\IEEEpeerreviewmaketitle

\section{Introduction}

\begin{figure}[!t]
\centering
\includegraphics[width=0.85\linewidth]{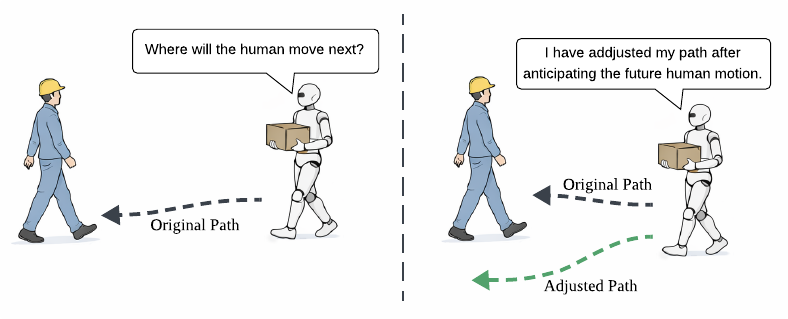}
\caption{\bluefunc{Illustration of human-humanoid robot interaction enabled by human motion prediction. The humanoid predicts the future motion of a human approaching on a collision path and adjusts its trajectory accordingly, enabling safe and efficient collaboration.}}
\label{fig:hrc_concept}
\end{figure}

Human motion prediction lies at the intersection of artificial intelligence, biomechanics, and robotics, aiming to forecast the complex, dynamic behaviors of people in real-world environments. This capability is pivotal for the next generation of intelligent systems, especially humanoid robots deployed in manufacturing~\cite{sajadieh2024towards, choi2025exploring}, healthcare~\cite{shanks2025cobotic}, and service domains~\cite{park2017disaster}. In industrial settings, for example, collaborative robots working alongside humans must anticipate a worker’s next move to coordinate seamlessly, avoid collisions, and maintain operational efficiency. Similarly, in healthcare and assisted living, predictive capabilities allow robotic aides to respond proactively to patient movements for timely intervention. Across these applications, accurate motion forecasting is a cornerstone of smooth and trustworthy Human-Robot Collaboration (HRC). \bluefunc{Recent works confirm that human motion prediction is a core component of such systems. Sampieri et al.~\cite{sampieri2022pose} trained pose forecasting models on Human3.6M~\cite{ionescu2013human3} and deployed them for collision detection in industrial cobot environments. Unhelkar et al.~\cite{unhelkar2018human} integrated motion prediction into robot planning for collaborative assembly. Together, these studies validate the pipeline of developing prediction modules on standard benchmarks and deploying them in manufacturing HRC. A conceptual illustration of this pipeline is shown in Fig.~\ref{fig:hrc_concept}.}

However, predicting human motion is inherently challenging due to its dependence on numerous variables such as individual physical characteristics, psychological state, task context, and environmental constraints~\cite{idrees2024human, lyu20223d}. These interwoven factors make it difficult to build models that generalize well across tasks and users, especially in dynamic and safety-critical domains like manufacturing. \bluefunc{From a physical AI perspective, human motion prediction plays a key role in enabling anticipatory behavior in embodied systems. In collaborative manufacturing environments, robots must continuously adapt their actions based on predicted human movements to ensure safety, efficiency, and smooth interaction~\cite{unhelkar2018human, liu2019deep, ding2024dynamic}.} The ability to generate predictions that are not only temporally accurate but also physically plausible is essential for enabling robots to operate safely and effectively in proximity to humans.

Early research on human motion prediction focused on sequential models such as Recurrent Neural Networks (RNNs) and their variants, which showed initial promise in modeling short sequences of human poses~\cite{erd, onhuman_rnn, gui2018adversarial, CHEN202476}. As the field progressed, researchers explored convolution-based approaches: Convolutional Neural Networks (CNNs) and graph-based models like Graph Convolutional Networks (GCNs) became popular for capturing spatial relationships between human body joints. These models improved pose representation by learning the human skeletal structure connectivity~\cite{li2018convolutional, liu2020trajectorycnn, dct1, mao2020history}. Additionally, generative models were introduced to better handle the inherent uncertainty in future motions. Approaches based on Generative Adversarial Networks (GANs) and Variational Autoencoders (VAEs) began to address issues like unrealistic or static predictions. Adversarial training, in particular, showed effectiveness in reducing implausible outcomes such as converging to mean poses or the zero-velocity problem, where the predictor yields frozen poses to minimize error~\cite{barsoum2018hp, kundu2019bihmp, chopin2021human}. Despite these advances, achieving accurate and natural long-range motion predictions remains an open problem. Many models struggle with compounding errors over time, often violating kinematic constraints (e.g., inconsistent limb lengths) or producing motion that, while minimizing short-term error, does not look realistic in a sustained sequence.

In recent years, the introduction of the Transformer network~\cite{transformer} has brought a paradigm shift in sequence modeling, including human motion prediction. Transformers, with their self-attention mechanism, excel at capturing dependencies in sequential data. Prior works have applied Transformers to human motion data, demonstrating improved ability to model both the temporal and spatial relationships in pose sequences~\cite{aksan2020attention, cai2020learning, zhao2023bidirectional, martinez2021pose}. These approaches significantly improve short-term human motion predictions. Nonetheless, challenges remain in leveraging Transformers for long-horizon motion forecasting. In particular, they often struggle to incorporate structural priors or capture inherent symmetries in human motion, typically requiring large-scale datasets or extensive data augmentation to generalize effectively~\cite{seo2025se}. As a result, ensuring physical realism and continuity over long durations demands additional innovation, since purely data-driven models can still drift into unnatural motion without appropriate constraints or feedback.

\begin{figure}[!t]
    \centering
    \includegraphics[width=\linewidth]{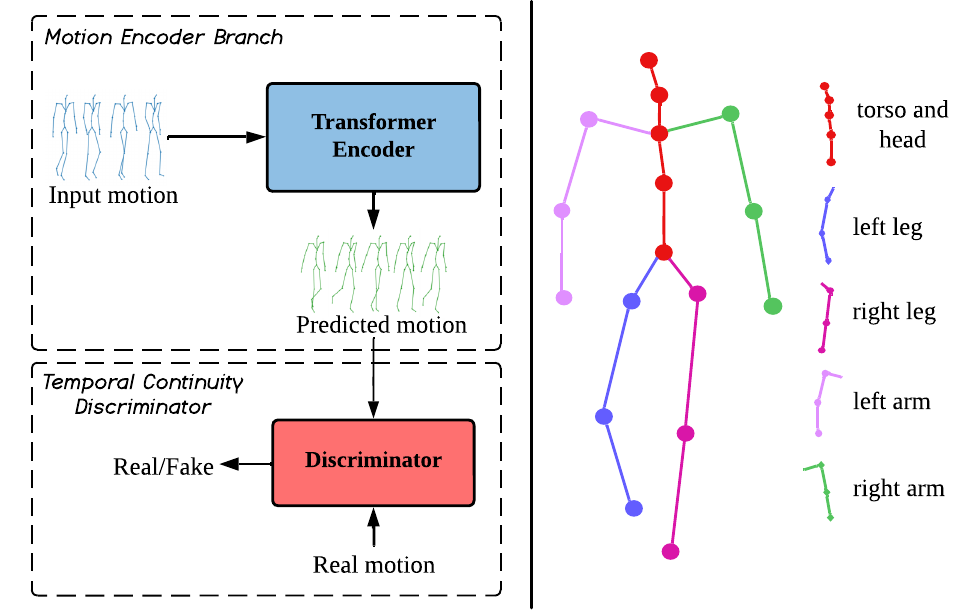}
    \caption[Long caption]{\textit{Left}: Overview of our proposed AdvMT network to predict future human motion by observing historical motion. \textit{Right}: The human body joints link structure consisting of human body parts: the torso and head, left leg, right leg, left arm, and right arm.}
    \label{arch_small}
\end{figure}

To address these challenges, we introduce Adversarial Motion Transformer (AdvMT), a novel architecture (illustrated in Fig.~\ref{arch_small}) that closes the gap between accuracy and realism in long-term human motion prediction. AdvMT combines the sequence modeling strength of a Transformer-based motion encoder with the physical plausibility constraints enforced through adversarial training. The encoder learns rich spatio-temporal features from past pose sequences, while a temporal continuity discriminator evaluates the realism of generated motion, guiding the model to produce human-like trajectories. A specialized loss function incorporating joint-level error, bone-length consistency, and adversarial feedback ensures the predicted poses adhere to human anatomical constraints.

Our autoregressive prediction framework allows past predicted frames to recursively inform future ones, closely simulating deployment scenarios in real-time robotic systems. \bluefunc{This approach targets the prediction capabilities that manufacturing environments require, where robots must anticipate human motion for safety and coordination~\cite{liu2019deep, ding2024dynamic}. Following the established practice of developing prediction modules on standard benchmarks before HRC deployment~\cite{sampieri2022pose}, we evaluate AdvMT on Human3.6M~\cite{ionescu2013human3}. The model matches or exceeds existing methods in short-term prediction and substantially improves long-term forecasts, providing a stronger basis for integration into collaborative robotic systems.}

\section{Related work}
\begin{figure*}[!t]
    \centering
    \includegraphics[width=0.85\linewidth]{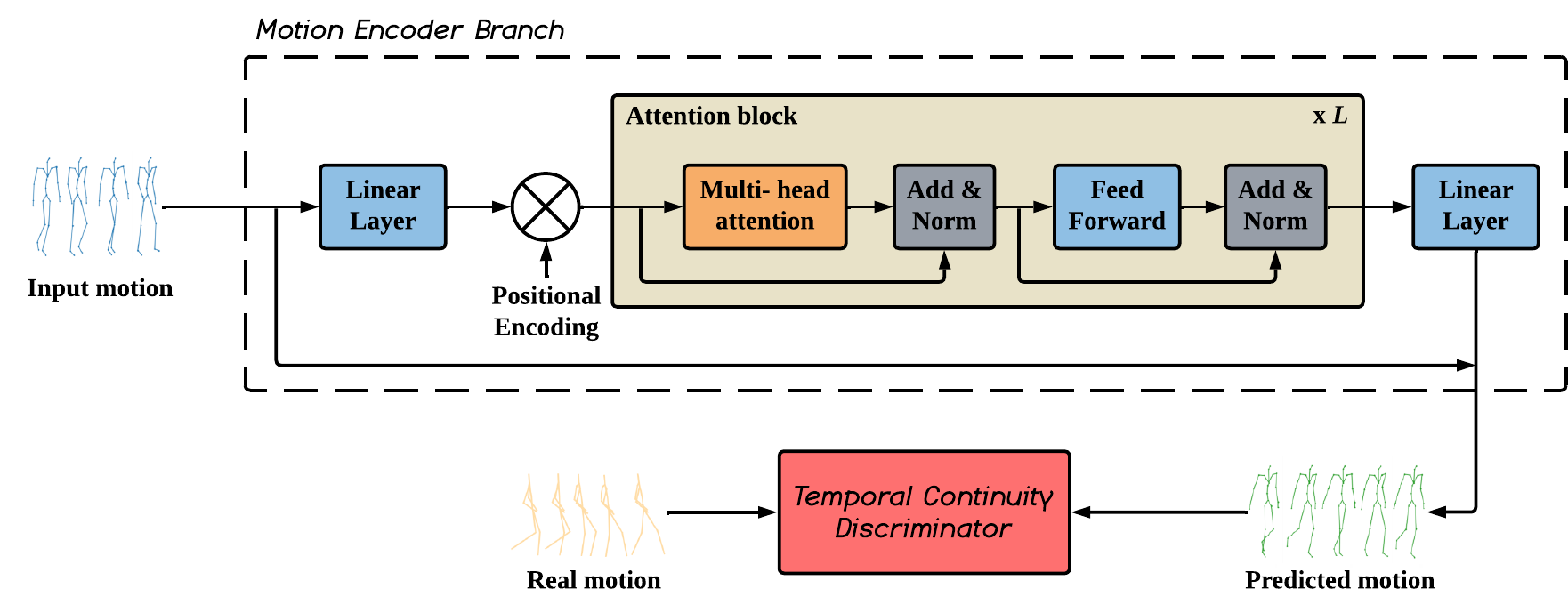}
 \caption[Long caption]{Overall AdvMT architecture with dual branches. The motion encoder branch is a Transformer-based network that encodes historical human motion and generates future pose predictions. The temporal continuity discriminator observes sequences of predicted poses and outputs a realism score. During training, the discriminator’s feedback guides the encoder to produce smooth, human-like motion transitions. This two-branch adversarial design ensures the predicted motion is both accurate and physically natural.}
    \label{overview}
   \end{figure*}
   
\subsection{Long-term human motion prediction}
The initial phase of research in the area of human motion prediction primarily utilized RNNs with encoder-decoder models~\cite{erd,onhuman_rnn, gui2018adversarial}, but these models faced challenges in handling complex human motion dependencies. This led to a shift towards CNNs, which provided improved extraction of spatial joint connection information~\cite{li2018convolutional, li2019efficient, liu2020trajectorycnn}. Later, the exploration of GCNs brought enhanced capabilities in anatomical relationship modeling~\cite{dct1, mao2020history, cui2020learning}. However, much of the research has concentrated mainly on short-term predictions, leaving a notable gap in long-term prediction accuracy.

While the field has made advancements in short-term motion prediction, efforts to address the extended horizon prediction challenge have been limited. Tang et al.~\cite{tang2018long} spearheaded this effort by effectively removing static joints, thereby focusing on joints in motion to predict reliable long-term motion.
To further advance this field, Xu et al.~\cite{xu2019ean} developed an attention-based Error Attenuation Network (EAN) with a focus on reducing error accumulation in future frames.

Furthermore, to leverage other architectures, Zhao et al.~\cite{zhao2023bidirectional} introduced a novel approach by integrating a Transformer network with the capabilities of GAN for long-term prediction challenges. 
Their method includes a bi-directional Transformer and both frame-level and sequence-level discriminators. 
While their model achieved enhanced accuracy for long-term predictions, this was accompanied by increased computational demands. In contrast, our approach outperforms this by intelligently combining a curated loss function and a temporal discriminator, resulting in more efficient predictions with a reduced parameter footprint.

\subsection{Adversarial training}
Traditional models in human motion prediction often struggled with ensuring motion smoothness and robustness. This challenge prompted researchers to investigate generative architectures and the potential of adversarial training. A significant development in this area was the AGED architecture~\cite{gui2018adversarial}, which utilized a geometry-aware adversarial learning technique. This approach not only improved motion cohesiveness but also introduced a new level of diversity in predictions. Following this, the Q-DCRN model~\cite{men2020quadruple} advanced the use of adversarial learning by employing a discriminator to refine motion predictions across various horizons.

The adoption of GANs in human motion prediction rapidly expanded, with studies~\cite{barsoum2018hp, kundu2019bihmp, hernandez2019human} exploring their capabilities. Despite their potential, GANs encountered challenges, especially in achieving Nash equilibrium. In response, refinement modules were integrated, as seen in~\cite{chao2020adversarial}, enhancing the accuracy of generated poses and minimizing artifacts. Building on these developments, Lyu et al.~\cite{lyu2021learning} took a novel approach by modeling joint motion through stochastic differential equations and utilized GANs for simulating path integrals, further refining the precision of human motion prediction.

\subsection{Transformer network}
The advent of the Transformer network marked a significant paradigm shift in sequence modeling~\cite{transformer}, offering a solution to the limitations inherent in RNNs, especially when dealing with long sequences. With their attention mechanisms, Transformers excel in focusing on pertinent features, thus efficiently handling long-term dependencies.  
\bluefunc{In human motion prediction, several works have applied Transformers with different architectural strategies. Aksan et al.~\cite{aksan2020attention} proposed a spatio-temporal Transformer that decouples attention into separate spatial and temporal blocks within each layer, where the spatial block attends across joints at a single timestep and the temporal block attends across timesteps for each individual joint. While this decoupled design captures inter-joint coordination and per-joint trajectories independently, it relies solely on L2 reconstruction loss with scheduled sampling and includes no mechanism to enforce physical plausibility or penalize unrealistic temporal transitions. Chen et al.~\cite{chen2022sttg} adopted a two-stage hybrid approach, combining a temporal Transformer with a spatial GCN and operating in the Discrete Cosine Transform (DCT) coefficient space. Although this pipeline achieves competitive short-term accuracy, its evaluation is limited to prediction horizons up to 400\,ms, and it similarly lacks adversarial feedback or biomechanical constraints.}

\bluefunc{In contrast to these approaches, our AdvMT employs an encoder-only Transformer with standard unified multi-head self-attention, which jointly captures both spatial and temporal dependencies within a single attention mechanism without requiring separate spatial modules such as a dedicated attention block or GCN. Our ablation study confirms that this simplified encoder-only design outperforms the full encoder-decoder Transformer for motion prediction. More importantly, AdvMT introduces two components absent from existing Transformer-based methods: (1)~a temporal continuity discriminator that provides adversarial feedback to enforce temporally realistic motion and prevent zero-velocity collapse, and (2)~a bone-length consistency loss that enforces anatomical constraints across predicted frames. This combination of architectural simplicity with adversarial training and physical priors is what enables AdvMT to achieve improved long-term prediction accuracy.}

\section{Adversarial Motion Transformer (AdvMT)}
The overall architecture of the Adversarial Motion Transformer (AdvMT), a system composed of two main branches, is illustrated in Fig.~\ref{overview}. The motion encoder branch interprets the input motion history and encodes the local and global human joint dependencies. The discriminator branch complements this by refining the predictions from the motion encoder branch to ensure the generation of realistic and consistent human motion.

% Results table (1st)
\begin{table*}[!t]
\centering
\caption{Comparison for short-term ($\le$400\,ms) and long-term ($>$400\,ms) prediction on Human3.6M~\cite{ionescu2013human3} across four action categories. \bluefunc{The symbols $*$, $\dagger$, and $\ddagger$ indicate that the corresponding results are taken from~\cite{mao2020history}, \cite{zhao2023bidirectional}, and \cite{guo2023back}, respectively.}}\label{table1}
\small\setlength{\tabcolsep}{4pt}
\begin{tabular}{@{}l@{\hspace{0.8em}}*{6}{r}@{\hspace{1.2em}}*{6}{r}@{}}
\toprule
& \multicolumn{6}{c}{\textbf{Walking}} & \multicolumn{6}{c}{\textbf{Eating}} \\
\cmidrule(lr){2-7}\cmidrule(l){8-13}
Time (ms) &160 &400 &560 &720 &880 &1000 &160 &400 &560 &720 &880 &1000 \\
\midrule
Res. Sup.$*$~\cite{onhuman_rnn} &40.9 &66.1 &71.6 &72.5 &76.0 &79.1 &31.5 &61.7 &74.9 &85.9 &93.8 &98.0 \\
convSeq2Seq$*$~\cite{li2018convolutional} &33.5 &63.6 &72.2 &77.2 &80.9 &82.3 &22.4 &48.4 &61.3 &72.8 &81.8 &87.1 \\
HisRepeat$*$~\cite{mao2020history} &\textbf{19.5} &39.8 &47.4 &52.1 &55.5 &58.1 &\textbf{14.0} &36.2 &50.0 &61.4 &70.6 &75.7 \\
BiTGAN$\dagger$~\cite{zhao2023bidirectional} &-- &-- &49.8 &55.0 &58.5 &60.5 &-- &-- &48.5 &59.2 &68.2 &73.0 \\
siMLPe$\ddagger$~\cite{guo2023back} &-- &\textbf{39.6} &46.8 &-- &-- &55.7 &-- &\textbf{36.1} &49.6 &-- &-- &74.5 \\
\midrule
AdvMT (ours) &23.9 &39.9 &\textbf{45.1} &\textbf{49.2} &\textbf{52.0} &\textbf{55.0} &18.3 &\textbf{36.1} &\textbf{44.6} &\textbf{51.5} &\textbf{56.5} &\textbf{59.3} \\
\midrule
& \multicolumn{6}{c}{\textbf{Smoking}} & \multicolumn{6}{c}{\textbf{Discussion}} \\
\cmidrule(lr){2-7}\cmidrule(l){8-13}
Time (ms) &160 &400 &560 &720 &880 &1000 &160 &400 &560 &720 &880 &1000 \\
\midrule
Res. Sup.$*$~\cite{onhuman_rnn} &34.7 &65.4 &78.1 &88.6 &96.6 &102.1 &47.8 &91.3 &109.5 &122.0 &128.6 &131.8 \\
convSeq2Seq$*$~\cite{li2018convolutional} &22.8 &48.9 &60.0 &69.4 &77.2 &81.7 &34.5 &77.6 &98.1 &112.9 &123.0 &129.3 \\
HisRepeat$*$~\cite{mao2020history} &\textbf{14.9} &36.4 &47.6 &\textbf{56.6} &\textbf{64.4} &69.5 &\textbf{23.4} &65.4 &86.6 &102.2 &113.2 &119.8 \\
BiTGAN$\dagger$~\cite{zhao2023bidirectional} &-- &-- &48.4 &57.5 &65.0 &70.0 &-- &-- &85.8 &101.2 &111.6 &116.4 \\
siMLPe$\ddagger$~\cite{guo2023back} &-- &\textbf{36.3} &\textbf{47.2} &-- &-- &\textbf{69.3} &-- &\textbf{64.3} &85.7 &-- &-- &116.3 \\
\midrule
AdvMT (ours) &22.8 &45.5 &56.5 &65.5 &72.7 &77.7 &36.0 &66.7 &\textbf{80.2} &\textbf{90.9} &\textbf{97.8} &\textbf{101.0} \\
\bottomrule
\end{tabular}
\end{table*}

\subsection{Problem formulation}
Human motion prediction fundamentally involves forecasting future movements by interpreting past sequences of human motion data. In a mathematical context, this can be visualized as a function that processes a series of historical human motion data points \( X_{1:T} = \{x_1, x_2, \dots, x_T\} \) and predicts future human poses. Each \( x_t = \{j_1, j_2, \dots, j_N\} \) in the sequence represents a single pose at time \( t \), consisting of \( N \) distinct joints. These joints are characterized in a \( K \)-dimensional pose representation, where \( K=3 \) signifies the 3D position representation in Euclidean space. The model is trained to predict the poses for the forthcoming \( L \) time steps, effectively forecasting the sequence \( \hat{X}_{T+1:T+L} \) based on the observed historical frames.

\subsection{Motion encoder branch}
The motion encoder branch in our model is based on the Transformer architecture, as described in the foundational work of Vaswani et al.~\cite{transformer}. Our implementation, however, deviates from the conventional Transformer framework by exclusively utilizing the encoder component. This specific design choice is grounded in our findings from extensive ablation studies, which demonstrate that the encoder alone is sufficient for capturing the complexities of human motion prediction. These studies revealed that employing the full Transformer architecture, including both encoder and decoder, tends to introduce unnecessary complexity without proportional benefits in this context. As a result, our focused approach with just the encoder component not only simplifies the model but also effectively enhances its capability to predict accurate human movements over extended timeframes.

\begin{figure}[!t]
 \centering
 \includegraphics[width=1\linewidth]{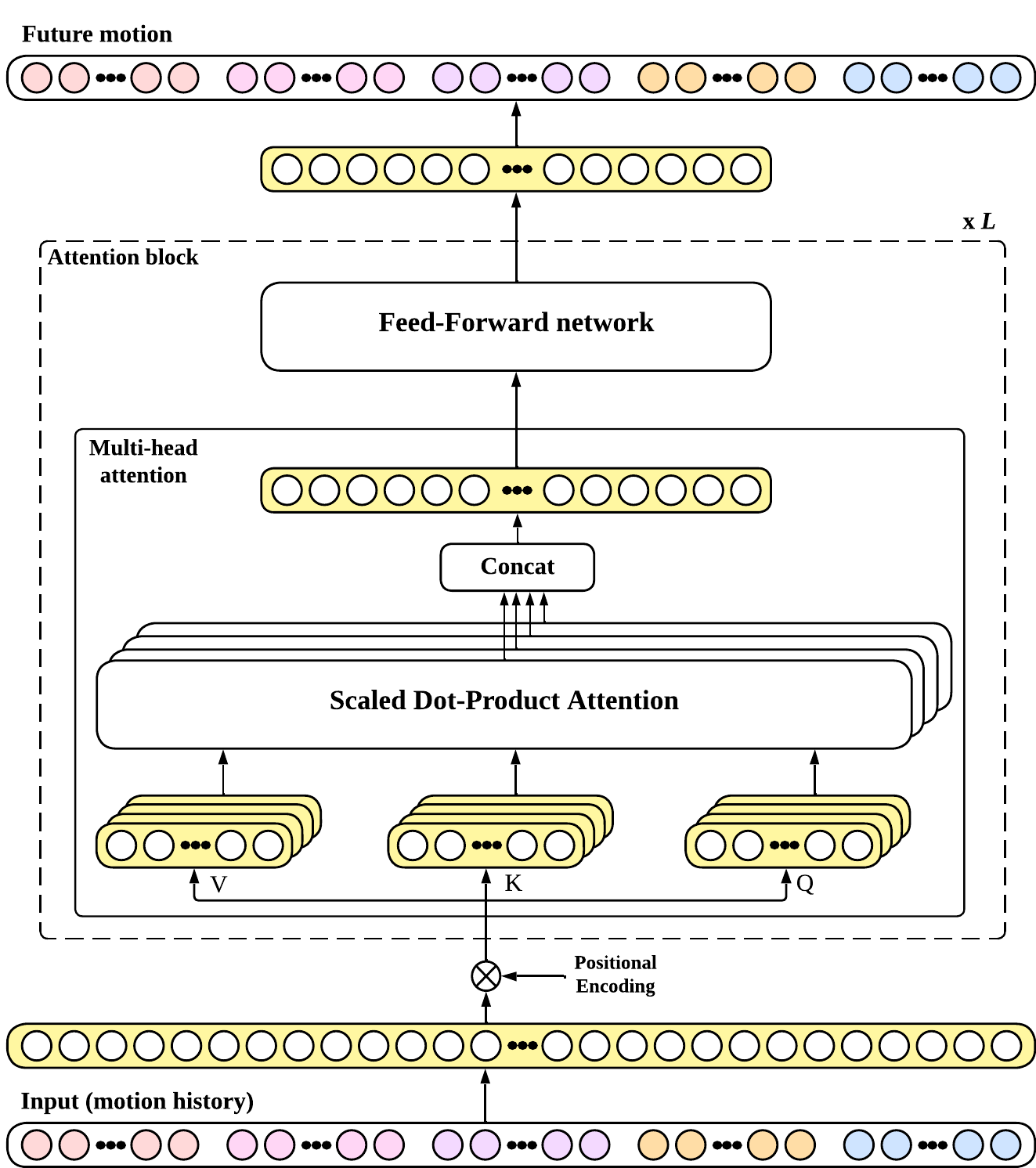}
 \caption{The detailed architecture of our Transformer-based motion encoder branch. The local and global dependencies within the human body are extracted through multiple layers of attention blocks. Each block aims to learn different aspects of motion dynamics, enabling a comprehensive understanding of human movement.}
 \label{motionencoder}
\end{figure}

A schematic of the motion encoder is depicted in Fig.~\ref{motionencoder}. The process is initiated by transforming the input pose data into joint embeddings through a linear layer. In line with the Transformer framework~\cite{transformer}, sinusoidal positional encodings are introduced to these embeddings. This addition is crucial as it allows the model to effectively process sequences of increased length, enabling a better understanding of the temporal relationships and positional dynamics within the motion data.

The architecture of our motion encoder comprises $L$ layers of attention blocks. Each block consists of a multi-head attention mechanism coupled with a position-wise feed-forward network. This configuration allows the model to concurrently learn and integrate various local and global dependencies present in the data. The aggregated representation, forged through these attention layers, is then projected back into the space of human poses through another linear layer. In the final stage of the process, the predicted future human pose is fed into a discriminator. This step is significant as it constrains the motion encoder to focus on learning patterns that result in realistic human motion.

\subsection{Temporal continuity discriminator}
Our primary goal in this research is to learn a robust and plausible representation of long-term human motion. Recognizing the inherent uncertainty in human behavior and actions, we observe that some models may overlook human body constraints and fail to predict human-like motion. To incorporate temporal continuity and human body constraints, we include a discriminator branch alongside the motion encoder branch. Our network is specifically trained to focus on the joint positions with an emphasis on maintaining natural body-joint velocities through adversarial learning. The key intuition here is to concentrate on the temporal differences in joint positions rather than their absolute values, ensuring a more continuous and realistic joint movement sequence. The adversarial loss of temporal continuity discriminator \(D_K\) is defined as,
\begin{equation}
\begin{split}
\mathcal{L}_{D_K} = \sum_{t=T+1}^{T+L} \left( \right. & \mathbb{E}_{x_t} \left[ \| D_K(\Delta x_t) \|^2 \right]  \\
& \left. + \mathbb{E}_{\hat{x}_t} \left[ \| 1 - D_K(\Delta \hat{x}_t) \|^2 \right] \right) ,
\end{split}
\end{equation}
where \(x_t\) and \(\hat{x}_t\) refers to real and predicted motion sequences respectively, and \(\Delta x\) is the temporal change in the motion sequence. \bluefunc{Rather than operating on absolute poses, the discriminator is applied to temporal pose differences computed with a fixed frame offset of 5, such that $\Delta x_t = x_t - x_{t-5}$. This design allows the discriminator to evaluate whether the generated temporal dynamics resemble those of real human motion.}

The autoregressive training regime empowers the discriminator to act as a feedback mechanism for the motion encoder branch, aiding in reducing error accumulation during extended horizon predictions. This approach significantly enhances motion prediction by preventing the tendency to predict zero-velocity motion. The inclusion of the discriminator branch not only enhances the realism of the generated motion but also ensures its smoothness over time.

% Results table (2nd)
\begin{table*}[!t]
\caption{Mean Per Joint Position Error (MPJPE) for long-term prediction ($>$400\,ms) on Human3.6M~\cite{ionescu2013human3} across remaining action categories. \bluefunc{The symbols $*$, $\dagger$, and $\ddagger$ indicate that the corresponding results are taken from~\cite{mao2020history}, \cite{zhao2023bidirectional}, and \cite{guo2023back}, respectively.}}\label{big_table}
\centering
\small\setlength{\tabcolsep}{4pt}
\begin{tabular}{@{}l@{\hspace{0.8em}}*{4}{r}@{\hspace{1.2em}}*{4}{r}@{\hspace{1.2em}}*{4}{r}@{}}
\toprule
& \multicolumn{4}{c}{\textbf{Directions}} & \multicolumn{4}{c}{\textbf{Greeting}} & \multicolumn{4}{c}{\textbf{Phoning}} \\
\cmidrule(lr){2-5}\cmidrule(lr){6-9}\cmidrule(l){10-13}
Time (ms) & 560 & 720 & 880 & 1000 & 560 & 720 & 880 & 1000 & 560 & 720 & 880 & 1000 \\
\midrule
Res. Sup.$*$~\cite{onhuman_rnn} & 101.1 & 114.5 & 124.5 & 129.1 & 126.1 & 138.8 & 150.3 & 153.9 & 94.0 & 107.7 & 119.1 & 126.4 \\
HisRepeat$*$~\cite{mao2020history} & 73.8 & 88.1 & 100.1 & 106.4 & 101.9 & 118.4 & 132.7 & 138.8 & 67.4 & 82.9 & 96.5 & 105.0 \\
BiTGAN$\dagger$~\cite{zhao2023bidirectional} & 73.3 & \textbf{87.9} & 99.7 & 106.3 & 101.1 & 117.8 & 131.4 & 136.4 & 67.3 & 82.3 & 94.9 & 103.2 \\
siMLPe$\ddagger$~\cite{guo2023back} & \textbf{73.1} & -- & -- & 106.7 & 99.8 & -- & -- & 137.5 & \textbf{66.3} & -- & -- & 103.3 \\
\midrule
AdvMT (ours) & 79.2 & 90.1 & \textbf{99.4} & \textbf{103.5} & \textbf{95.1} & \textbf{104.5} & \textbf{114.1} & \textbf{118.5} & 68.4 & \textbf{79.6} & \textbf{88.5} & \textbf{93.7} \\
\midrule
& \multicolumn{4}{c}{\textbf{Posing}} & \multicolumn{4}{c}{\textbf{Purchases}} & \multicolumn{4}{c}{\textbf{Sitting}} \\
\cmidrule(lr){2-5}\cmidrule(lr){6-9}\cmidrule(l){10-13}
Time (ms) & 560 & 720 & 880 & 1000 & 560 & 720 & 880 & 1000 & 560 & 720 & 880 & 1000 \\
\midrule
Res. Sup.$*$~\cite{onhuman_rnn} & 140.3 & 159.8 & 173.2 & 183.2 & 122.1 & 137.2 & 148.0 & 154.0 & 113.7 & 130.5 & 144.4 & 152.6 \\
HisRepeat$*$~\cite{mao2020history} & 107.5 & 136.8 & 161.4 & 178.2 & 95.5 & \textbf{110.9} & 125.0 & 134.2 & 76.4 & 93.1 & 107.0 & 116.0 \\
BiTGAN$\dagger$~\cite{zhao2023bidirectional} & 107.1 & 134.6 & 156.7 & 171.0 & 99.0 & 113.7 & 127.1 & 135.1 & 76.0 & \textbf{92.0} & \textbf{105.4} & 114.4 \\
siMLPe$\ddagger$~\cite{guo2023back} & \textbf{103.4} & -- & -- & 168.7 & \textbf{93.8} & -- & -- & 132.5 & \textbf{75.4} & -- & -- & \textbf{114.1} \\
\midrule
AdvMT (ours) & 114.1 & \textbf{128.1} & \textbf{138.4} & \textbf{145.2} & 99.2 & 112.3 & \textbf{121.8} & \textbf{127.9} & 87.2 & 101.8 & 113.1 & 121.4 \\
\midrule
& \multicolumn{4}{c}{\textbf{Sitting Down}} & \multicolumn{4}{c}{\textbf{Taking Photo}} & \multicolumn{4}{c}{\textbf{Waiting}} \\
\cmidrule(lr){2-5}\cmidrule(lr){6-9}\cmidrule(l){10-13}
Time (ms) & 560 & 720 & 880 & 1000 & 560 & 720 & 880 & 1000 & 560 & 720 & 880 & 1000 \\
\midrule
Res. Sup.$*$~\cite{onhuman_rnn} & 138.8 & 159.0 & 176.1 & 187.4 & 110.6 & 128.9 & 143.7 & 153.9 & 105.4 & 117.3 & 128.1 & 135.4 \\
HisRepeat$*$~\cite{mao2020history} & 97.0 & 116.1 & 132.1 & 143.5 & 72.1 & \textbf{90.0} & \textbf{105.5} & 115.9 & 74.5 & 89.0 & 100.3 & 108.2 \\
BiTGAN$\dagger$~\cite{zhao2023bidirectional} & 96.2 & \textbf{114.5} & \textbf{129.9} & \textbf{141.3} & 74.2 & 92.6 & 107.4 & 117.7 & 72.9 & \textbf{87.3} & \textbf{97.7} & 104.9 \\
siMLPe$\ddagger$~\cite{guo2023back} & \textbf{95.7} & -- & -- & 142.4 & \textbf{71.0} & -- & -- & \textbf{112.8} & \textbf{71.6} & -- & -- & \textbf{104.6} \\
\midrule
AdvMT (ours) & 100.5 & 117.9 & 132.1 & 142.2 & 85.4 & 100.7 & 113.4 & 122.0 & 83.3 & 94.8 & 103.5 & 109.5 \\
\midrule
& \multicolumn{4}{c}{\textbf{Walking Dog}} & \multicolumn{4}{c}{\textbf{Walking Together}} & \multicolumn{4}{c}{\textbf{Average}} \\
\cmidrule(lr){2-5}\cmidrule(lr){6-9}\cmidrule(l){10-13}
Time (ms) & 560 & 720 & 880 & 1000 & 560 & 720 & 880 & 1000 & 560 & 720 & 880 & 1000 \\
\midrule
Res. Sup.$*$~\cite{onhuman_rnn} & 128.7 & 141.1 & 155.3 & 164.5 & 80.2 & 87.3 & 92.8 & 98.2 & 106.3 & 119.4 & 130.0 & 136.6 \\
HisRepeat$*$~\cite{mao2020history} & 108.2 & 120.6 & 135.9 & 146.9 & 52.7 & \textbf{57.8} & \textbf{62.0} & 64.9 & 77.3 & 91.8 & 104.1 & 112.1 \\
BiTGAN$\dagger$~\cite{zhao2023bidirectional} & 105.4 & 120.4 & 136.4 & 148.3 & 54.3 & 59.7 & 64.2 & 67.3 & 77.3 & 91.7 & 103.6 & 111.1 \\
siMLPe$\ddagger$~\cite{guo2023back} & 105.6 & -- & -- & 141.2 & \textbf{50.8} & -- & -- & \textbf{61.5} & \textbf{75.7} & -- & -- & 109.4 \\
\midrule
AdvMT (ours) & \textbf{102.5} & \textbf{115.5} & \textbf{127.7} & \textbf{136.8} & 62.9 & 71.0 & 79.0 & 84.4 & 80.3 & \textbf{91.6} & \textbf{100.8} & \textbf{106.6} \\
\bottomrule
\end{tabular}

\end{table*}

\subsection{Loss function}
Studies have shown that using a vanilla Euclidean loss in human motion prediction often causes models to converge to a mean pose, a phenomenon highlighted in~\cite{lyu20223d}. To address this issue, we propose a modified loss function designed to capture a better representation of motion data. Additionally, previous research has encountered challenges with zero-velocity collapse in long-term prediction tasks. This means that for predictions extending over 400\,ms, models often default to predicting static outputs for future frames, failing to capture the dynamic nature of human motion. Our modified loss function aims to mitigate these issues, enhancing the ability of the model to accurately predict longer sequences of human movement.

To effectively learn true human motion representation, we address both spatial and temporal dependencies in the input motion sequence. The spatial relationship between human joints within each frame is captured by the self-attention layer in our motion encoder branch. Simultaneously, we ensure temporal consistency across frames by integrating it into our loss function formulation. This approach is vital for accurate predictions over longer time horizons, embedding a comprehensive understanding of both spatial and temporal dynamics directly into the training regime. Our tailored loss function is defined as

\begin{equation}
\mathcal{L}(X, \hat{X}) = \mathcal{L}_{\text{MPJPE}} + \lambda_{B} \mathcal{L}_{\text{bone}} + \lambda_{D} \mathcal{L}_{D_K},
\end{equation}
where $X$ and $\hat{X}$ are ground truth and predicted poses. The first term in our loss function corresponds to the Mean Per Joint Position Error (MPJPE) as proposed in~\cite{ionescu2013human3}. With $t$ and $n$ as the frame and joint number respectively, the \(\mathcal{L}_{\text{MPJPE}}\) is defined as
\begin{equation}
    \mathcal{L}_{\text{MPJPE}} = \frac{1}{N(T + L)} \sum_{t=T+1}^{T + L} \sum_{n=1}^{N} \left\| \hat{x}_{t,n} - x_{t,n} \right\|^2.
\end{equation}

Moreover, the terms $\mathcal{L}_{\text{bone}}$ and $\mathcal{L}_{D_K}$ represent the bone length error and the adversarial loss, respectively. Each term is weighted by its regularization factor, with \( \lambda_{B} \) for bone length error and \( \lambda_{D} \) for adversarial loss, ensuring a balanced contribution of each component to the overall loss function. Our rationale for incorporating bone length error is based on the constant nature of bone lengths over time, which introduces temporal consistency into our motion predictions. Bone lengths are calculated as the Euclidean distance between connected joint positions in both predicted and ground truth poses.

However, relying solely on bone length error has a drawback. It may lead the motion encoder to minimize this error without adequately addressing zero-velocity, resulting in static motion predictions. To counteract this, we introduce the temporal continuity discriminator loss $\mathcal{L}_{D_K}$, which penalizes unrealistic human motions. This additional loss encourages the motion encoder to generate more dynamic, realistic, and plausible human motion, thus addressing a critical aspect of motion prediction that bone length error alone cannot resolve.

\section{Experiments}
\begin{figure}[!t]
 \centering
 \includegraphics[width=0.95\linewidth]{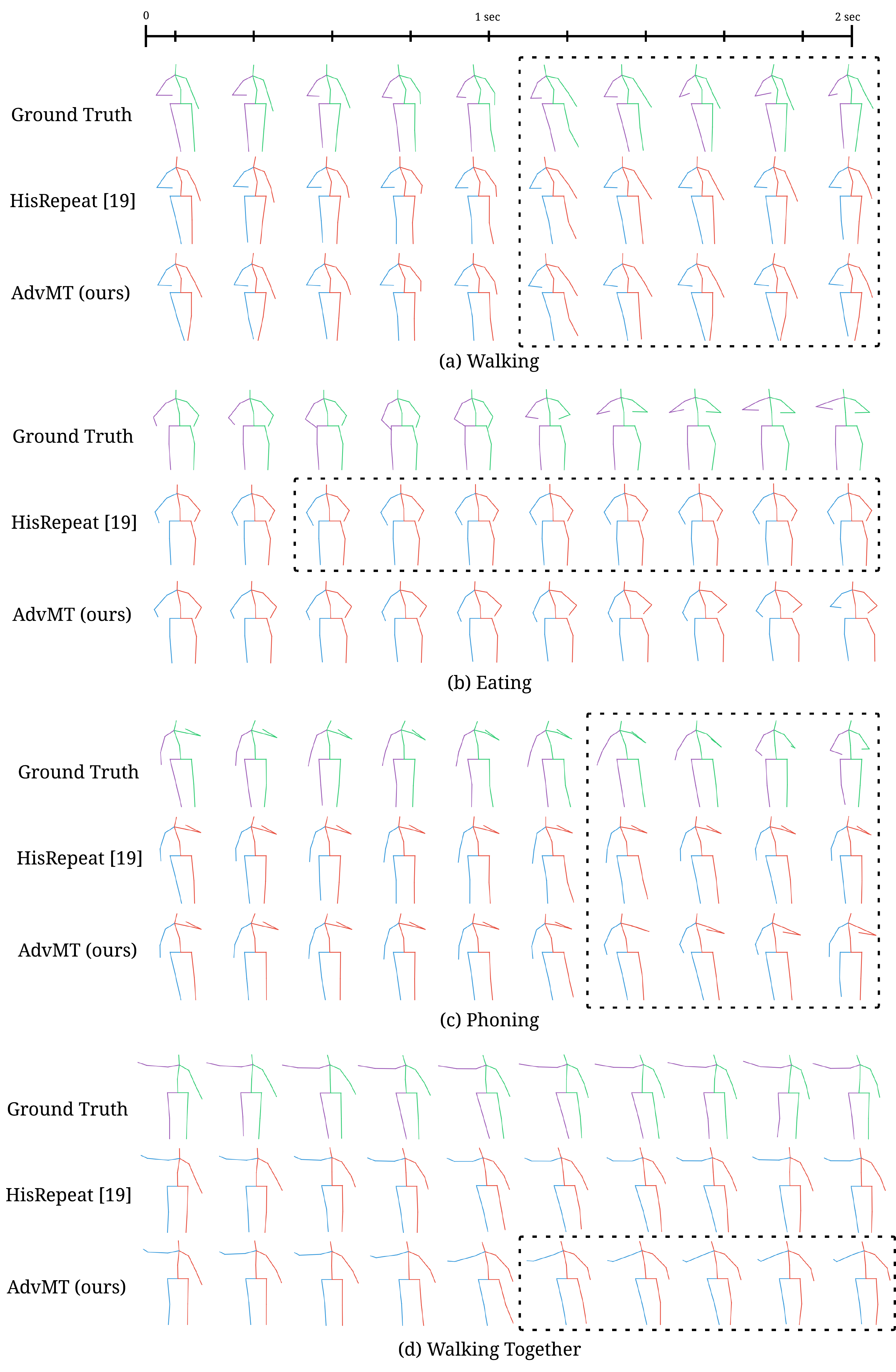}
 \caption[Long caption]{Qualitative results up to 2 seconds future motion prediction for \textit{walking}, \textit{eating}, \textit{phoning}, and \textit{walking together} actions from Human3.6M dataset~\cite{ionescu2013human3}. For visualization purposes, the predictions are downsampled to 5 frames per second. Ground truth poses are drawn in purple and green, whereas the future predictions are marked in blue and red colors. Best visualized in zoomed view.}
 \label{drawing}
\end{figure}

\subsection{Dataset} 
In our experiments, we conducted our model evaluation using the Human3.6M dataset~\cite{ionescu2013human3}, widely recognized as a benchmark in the field of human motion prediction. This extensive database contains over 3.6 million 3D poses, recorded with 7 actors performing 15 different types of actions. \bluefunc{Following prior work~\cite{onhuman_rnn,mao2020history,zhao2023bidirectional,guo2023back}, we downsample the motion sequences from 50 fps to 25 fps and remove the global rotation, global translation, and constant joint dimensions from each pose representation. In alignment with the training and evaluation protocol outlined in~\cite{onhuman_rnn,mao2020history,zhao2023bidirectional,guo2023back}, we used subjects S1, S6, S7, S8, S9, and S11 for training, and S5 is designated for evaluation, and report the results for 256 sub-sequences for each action.}

\bluefunc{\subsection{Implementation details}}

% \bluefunc{\subsubsection{Transformer motion encoder}}
\bluefunc{The motion encoder in AdvMT is implemented as an encoder-only Transformer with an autoregressive prediction strategy. The input 3D joint-position sequence is first projected into a 256-dimensional latent space through a linear layer, followed by positional encoding and layer normalization as explained in~\cite{transformer}. The encoder consists of four stacked Transformer blocks, where each block contains a 4-head self-attention module with $d_k=d_v=32$, followed by a position-wise feed-forward network with the input and output layer size of $256$, and inner-layer size of $512$ with ReLU activation. Similar to~\cite{transformer}, the residual connections and layer normalization are applied throughout the network, and a dropout rate of 0.3 is used. The final hidden representation is projected back to the pose space through a linear layer, and a residual skip connection is added between the input and output pose representation. During prediction, given 50 observed frames, the model generates future motion autoregressively by predicting one frame at a time, appending the predicted frame to the input window, discarding the oldest frame, and repeating this process until the full future sequence is generated.}

\bluefunc{The discriminator is implemented as a lightweight 1D convolutional network composed of two convolutional layers of size $32$x$32$ with ReLU activations, followed by adaptive average pooling and per-channel fully connected outputs. During training, we adopt a least-squares adversarial objective and optimize the generator and discriminator with alternating 1:1 updates. In this way, the discriminator complements the reconstruction objective by penalizing temporally implausible predictions and reducing the tendency toward static or zero-velocity motion.}

% \bluefunc{\subsubsection{Training settings}}
\bluefunc{The generator is optimized using AdamW with $\beta_1=0.9$, $\beta_2=0.999$, and weight decay of 0.01, while the discriminator is optimized using Adam with a fixed learning rate of $1\times10^{-4}$. For the motion encoder, we use the standard Transformer~\cite{transformer} warmup schedule with 10{,}000 warmup steps. Training is performed for 30 epochs with a batch size of 32. The loss weights are set to $\lambda_B = 0.3$ and $\lambda_D = 0.3$.}
\vspace{1.5em}

\subsection{Comparison with other methods}
In our study, we focused on training our method for 3D joint position prediction. We compared its performance with other state-of-the-art methods trained for the same motion representation, as detailed in Tables~\ref{table1} and~\ref{big_table}. The primary metric used for evaluation is the MPJPE in millimeters, aligning with the standards used by other methods~\cite{mao2020history, CAO2022106, zhao2023bidirectional, guo2023back}. We trained the model using an input sequence of 2 seconds to generate predictions for the next 1 second. However, our motion encoder branch was trained autoregressively, enabling it to predict motions extending beyond 1 second.

The results reveal that our proposed method consistently surpasses the baseline method~\cite{onhuman_rnn} in both short-term and long-term predictions. While our performance in short-term prediction is comparable to the current state-of-the-art, it is in long-term prediction where our method particularly excels, outperforming in most of the action tasks evaluated.

The qualitative results further substantiate the efficacy of our model. As demonstrated in Fig.~\ref{drawing}, our model excels in accurately predicting joint movements while adhering to the constraints of human body movement. In dynamic sequences like walking, our model significantly improves leg movement accuracy compared to the results in~\cite{mao2020history}. In addition, for actions like eating and phoning, our model closely approximates actual hand movements, unlike the static outputs predicted in the final frames by~\cite{mao2020history}.

Due to its autoregressive prediction approach, our method successfully avoids predicting zero-velocity motion for long-term predictions. A prime example is the phoning action, where our model realistically simulates the action of ending a call and putting down the phone, as marked with a dashed box in Fig.~\ref{drawing}. It is important to note that while our method excels in dynamic actions, its performance is slightly reduced in static actions such as smoking and waiting.

\bluefunc{To quantitatively validate the long-horizon prediction capability, we extend the evaluation to 2 seconds (50 frames). Table~\ref{table_extended} reports MPJPE for AdvMT and HisRepeat~\cite{mao2020history} from 1200\,ms to 2000\,ms across all 15 actions. HisRepeat~\cite{mao2020history} predictions were generated using the trained models provided by the authors. AdvMT maintains lower average error across the extended horizon on 14 of 15 actions.}

\begin{table*}[!t]
\caption{\bluefunc{Extended motion prediction until 2 seconds on Human3.6M~\cite{ionescu2013human3} across all 15 action categories. HisRepeat~\cite{mao2020history} predictions were generated using the trained models provided by the authors. Bold indicates lower error.}}\label{table_extended}
\centering
\footnotesize\setlength{\tabcolsep}{1.5pt}\renewcommand{\arraystretch}{1.02}
\begin{tabular}{@{}l@{\hspace{0.35em}}*{5}{r}@{\hspace{0.55em}}*{5}{r}@{\hspace{0.55em}}*{5}{r}@{\hspace{0.55em}}*{5}{r}@{}}
\toprule
& \multicolumn{5}{c}{\textbf{Walking}} & \multicolumn{5}{c}{\textbf{Eating}} & \multicolumn{5}{c}{\textbf{Smoking}} & \multicolumn{5}{c}{\textbf{Discussion}} \\
\cmidrule(lr){2-6}\cmidrule(lr){7-11}\cmidrule(lr){12-16}\cmidrule(l){17-21}
Time (ms) & 1200 & 1400 & 1600 & 1800 & 2000 & 1200 & 1400 & 1600 & 1800 & 2000 & 1200 & 1400 & 1600 & 1800 & 2000 & 1200 & 1400 & 1600 & 1800 & 2000 \\
\midrule
HisRepeat~\cite{mao2020history} & 59.1 & 61.6 & 66.5 & 72.2 & 73.4 & 82.6 & 87.9 & 90.7 & 93.1 & 96.1 & \textbf{77.2} & \textbf{84.0} & \textbf{89.4} & \textbf{94.8} & 101.8 & 131.3 & 138.6 & 144.3 & 149.0 & 151.1 \\
AdvMT (ours) & \textbf{56.3} & \textbf{59.7} & \textbf{65.4} & \textbf{71.0} & \textbf{73.2} & \textbf{63.6} & \textbf{66.6} & \textbf{68.7} & \textbf{70.5} & \textbf{71.4} & 84.1 & 89.4 & 92.2 & 95.1 & \textbf{99.1} & \textbf{106.3} & \textbf{109.4} & \textbf{114.1} & \textbf{117.7} & \textbf{121.1} \\
\midrule
& \multicolumn{5}{c}{\textbf{Directions}} & \multicolumn{5}{c}{\textbf{Greeting}} & \multicolumn{5}{c}{\textbf{Phoning}} & \multicolumn{5}{c}{\textbf{Posing}} \\
\cmidrule(lr){2-6}\cmidrule(lr){7-11}\cmidrule(lr){12-16}\cmidrule(l){17-21}
Time (ms) & 1200 & 1400 & 1600 & 1800 & 2000 & 1200 & 1400 & 1600 & 1800 & 2000 & 1200 & 1400 & 1600 & 1800 & 2000 & 1200 & 1400 & 1600 & 1800 & 2000 \\
\midrule
HisRepeat~\cite{mao2020history} & 116.3 & 121.4 & 126.7 & 130.0 & 132.9 & 148.2 & 152.4 & 153.1 & 150.2 & 150.6 & 118.9 & 132.4 & 144.2 & 153.4 & 162.3 & 202.5 & 220.4 & 233.3 & 246.4 & 254.6 \\
AdvMT (ours) & \textbf{109.2} & \textbf{112.4} & \textbf{115.1} & \textbf{117.8} & \textbf{119.2} & \textbf{125.8} & \textbf{129.4} & \textbf{129.0} & \textbf{126.9} & \textbf{126.5} & \textbf{101.2} & \textbf{108.0} & \textbf{114.6} & \textbf{119.4} & \textbf{125.1} & \textbf{155.9} & \textbf{163.9} & \textbf{170.4} & \textbf{177.9} & \textbf{183.8} \\
\midrule
& \multicolumn{5}{c}{\textbf{Purchases}} & \multicolumn{5}{c}{\textbf{Sitting}} & \multicolumn{5}{c}{\textbf{Sitting Down}} & \multicolumn{5}{c}{\textbf{Taking Photo}} \\
\cmidrule(lr){2-6}\cmidrule(lr){7-11}\cmidrule(lr){12-16}\cmidrule(l){17-21}
Time (ms) & 1200 & 1400 & 1600 & 1800 & 2000 & 1200 & 1400 & 1600 & 1800 & 2000 & 1200 & 1400 & 1600 & 1800 & 2000 & 1200 & 1400 & 1600 & 1800 & 2000 \\
\midrule
HisRepeat~\cite{mao2020history} & 145.9 & 154.0 & 162.5 & 166.2 & 172.2 & \textbf{132.3} & 147.7 & 158.5 & 168.2 & 178.9 & 159.9 & 177.6 & 191.5 & 204.3 & 217.4 & \textbf{133.1} & 148.5 & 161.0 & 172.3 & 182.3 \\
AdvMT (ours) & \textbf{134.7} & \textbf{137.0} & \textbf{142.7} & \textbf{146.9} & \textbf{153.6} & 134.9 & \textbf{146.2} & \textbf{153.3} & \textbf{157.7} & \textbf{162.9} & \textbf{157.6} & \textbf{171.3} & \textbf{181.2} & \textbf{188.8} & \textbf{195.6} & 134.3 & \textbf{145.8} & \textbf{153.6} & \textbf{161.0} & \textbf{167.9} \\
\midrule
& \multicolumn{5}{c}{\textbf{Waiting}} & \multicolumn{5}{c}{\textbf{Walking Dog}} & \multicolumn{5}{c}{\textbf{Walking Together}} & \multicolumn{5}{c}{\textbf{Average}} \\
\cmidrule(lr){2-6}\cmidrule(lr){7-11}\cmidrule(lr){12-16}\cmidrule(l){17-21}
Time (ms) & 1200 & 1400 & 1600 & 1800 & 2000 & 1200 & 1400 & 1600 & 1800 & 2000 & 1200 & 1400 & 1600 & 1800 & 2000 & 1200 & 1400 & 1600 & 1800 & 2000 \\
\midrule
HisRepeat~\cite{mao2020history} & 119.5 & 129.5 & 138.1 & 147.8 & 156.3 & 158.4 & 170.2 & 180.2 & 189.8 & 195.7 & \textbf{69.2} & \textbf{72.0} & \textbf{76.7} & \textbf{82.4} & \textbf{85.7} & 123.6 & 133.2 & 141.1 & 148.0 & 154.1 \\
AdvMT (ours) & \textbf{118.8} & \textbf{127.7} & \textbf{134.6} & \textbf{141.4} & \textbf{145.7} & \textbf{145.8} & \textbf{151.6} & \textbf{155.1} & \textbf{162.7} & \textbf{168.5} & 84.4 & 92.7 & 100.1 & 105.5 & 109.6 & \textbf{114.8} & \textbf{121.2} & \textbf{126.4} & \textbf{131.0} & \textbf{135.1} \\
\bottomrule
\end{tabular}
\end{table*}

% \bluefunc{These results highlight the advantages of AdvMT's design over existing Transformer-based methods. Prior approaches such as Aksan et al.~\cite{aksan2020attention}, which uses decoupled spatial-temporal attention with only reconstruction loss, and Chen et al.~\cite{chen2022sttg}, which combines a temporal Transformer with a spatial GCN trained with L1 loss, are both prone to error accumulation and zero-velocity convergence over extended horizons due to the absence of adversarial feedback. In contrast, AdvMT's unified encoder-only architecture, combined with the temporal continuity discriminator, actively penalizes static or temporally implausible predictions during training. The ablation study in Table~\ref{ablationTable} confirms the contribution of each component: compared to the encoder trained with only MPJPE loss, the addition of the discriminator and bone-length loss reduces the overall error. This demonstrates that adversarial training and physical constraints, absent in prior Transformer-based methods, are essential for accurate long-horizon prediction.}

% A limitation we observed in our method is its tendency to focus on specific parts of the human body when predicting future motion. For example, in the walking together action, which requires learning both lower and upper body movements, our method concentrates on lower body movements, resulting in less accurate predictions of hand positions.

\begin{table}[!t]
\caption{Ablation study results on the Human3.6M dataset~\cite{ionescu2013human3}.}\label{ablationTable}
\centering
\small\setlength{\tabcolsep}{3pt}\renewcommand{\arraystretch}{1}
\begin{tabular*}{\columnwidth}{@{}l@{\extracolsep{\fill}}ccccc@{}}
\toprule
\multirow{2}{*}{\textbf{Methods}} & \multicolumn{5}{c}{\textbf{Time (ms)}} \\
\cmidrule(lr){2-6}
& 160 & 400 & 560 & 880 & 1000 \\
\midrule
Baseline~\cite{transformer} & 44.2 & 79.7 & 92.2 & 118.9 & 126.6 \\
AdvMT ($\mathcal{L}_{\text{MPJPE}}$) & 45.8 & 77.2 & 88.9 & 112.9 & 119.7 \\
\shortstack[l]{AdvMT ($\mathcal{L}_{\text{MPJPE}}$ +\\$\mathcal{L}_{\text{bone}}$ + $\mathcal{L}_{D_K}$)} & \textbf{33.2} & \textbf{65.3} & \textbf{80.3} & \textbf{100.8} & \textbf{106.6} \\
\bottomrule
\end{tabular*}
\end{table}

\section{Ablation study}

\subsection{Architecture}
An ablation study was conducted to assess the effectiveness of our proposed AdvMT architecture. We compared the performance of a full Transformer network against our modified architecture. Originally intended for sequence-to-sequence tasks, the full Transformer architecture was found less effective for human motion prediction compared to our adaptation, which solely utilizes the encoder layer. This suggests that the decoder layer might add unnecessary complexity, hindering the ability to capture human motion dynamics accurately.

Further ablation studies highlighted the critical role of the discriminator branch in enhancing the realism and temporal consistency of generated motion. The discriminator branch functions as a \textit{critic}, ensuring the predicted motions are not only realistic but also align with real-world motion patterns. Its absence leads to the generation of unrealistic or inconsistent motion sequences, particularly in complex human motions where accuracy in joint angles and movements is crucial. By integrating the adversarial training, our model is compelled to generate more lifelike and consistent motion sequences.

\subsection{Loss function}
Regarding the loss function, we found that using only the bone length error as a loss term with the vanilla MPJPE loss led the model to predict zero-velocity motion for long-term predictions. As the model learns to minimize only the bone length error, leading it to generate static poses. To mitigate this, incorporating the discriminator loss term helped to penalize unrealistic human motion predictions and encouraged the motion encoder to generate more realistic and plausible human motion. We conducted an ablation study to investigate the effectiveness of our modified loss function, which includes a discriminator loss and a bone error loss in addition to the regular MPJPE loss. We compared the performance of our model with the modified loss function against a baseline model that only used the regular MPJPE loss. The results show that our modified loss function, which includes the MPJPE loss, bone length error, and a temporal continuity discriminator loss, is effective in improving the quality of human motion prediction.

\bluefunc{\subsection{Discussion}}

\bluefunc{An analysis of the per-action results in Tables~\ref{table1} and~\ref{big_table} reveals distinct
performance patterns. AdvMT's improvements are most pronounced for dynamic actions involving
substantial joint movement in walking, greeting, and walking dog actions, where the temporal continuity
discriminator prevents zero-velocity convergence. For static actions, such as smoking, sitting, and 
waiting, the advantage is reduced because the ground truth exhibits minimal temporal
variation. Nonetheless, the overall average MPJPE across all actions is the lowest
among all compared methods.}

\bluefunc{We also observe a trade-off between short-term and long-term accuracy. At
short-term predictions, AdvMT's error is comparable to existing methods, whereas at longer horizons, it
consistently achieves the best performance. The adversarial training objective guides the model to
prioritize temporally coherent trajectories over frame-level positional accuracy, distributing
prediction quality across the full horizon. For human-robot collaboration, where reliable
motion forecasts several hundred milliseconds ahead are needed for proactive planning, this
trade-off is desirable.}

\bluefunc{Finally, in multi-part actions such as walking together, the attention
mechanism tends to concentrate on the most kinematically active body region, i.e., lower body, while
producing less accurate predictions for the upper body. This is consistent with the per-channel discriminator design, since it operates
on unnormalized temporal differences, joints with larger displacements 
produce larger convolutional activations and stronger gradients to the generator, while joints
with smaller motion receive weaker supervision. Future work
could address this through structure-aware attention, multi-scale discriminators for different
body regions, or explicit kinematic chain constraints.}

\section{Conclusion}

In this work, we introduced AdvMT, a Transformer-based architecture designed for long-term human motion prediction with an emphasis on temporal consistency and physical plausibility. By combining a spatially attentive motion encoder with a tailored loss function and a temporal continuity discriminator, AdvMT effectively mitigates prediction drift and enforces smooth transitions across frames. Our autoregressive training regime, guided by adversarial feedback, enables accurate short-term predictions while significantly improving long-horizon forecasts across diverse action classes. These advancements are particularly relevant for human-robot collaboration in manufacturing environments, where anticipating human movement is essential for safety and coordination. AdvMT’s ability to generate biomechanically realistic motion trajectories can facilitate more seamless and intelligent interaction between robots and human workers. Future extensions may incorporate structure-aware encoders or explicit kinematic constraints to enhance performance in early prediction intervals and further support generalization to complex real-world tasks.

%\section*{Acknowledgment}

\ifCLASSOPTIONcaptionsoff
  \newpage
\fi



%\bibliographystyle{IEEEtran}
\bibliographystyle{ieeetr}
\bibliography{ref}

\end{document}